\pdfoutput=1

\documentclass[11pt]{article}

\usepackage[preprint]{acl}

\usepackage{times}
\usepackage{latexsym}

\usepackage[utf8]{inputenc}
\usepackage{forest}
\usepackage{hyperref}
\usepackage{amsmath}
\usepackage{subcaption}
\usepackage{geometry}
\usepackage{amsfonts}
\usepackage[justification=raggedright,singlelinecheck=false]{caption}

\usepackage[T1]{fontenc}

\usepackage[utf8]{inputenc}

\usepackage{microtype}

\usepackage{inconsolata}

\usepackage{graphicx}

\title{Thinking Machines: A Survey of LLM based Reasoning Strategies}
 
\author{Dibyanayan Bandyopadhyay$^1$, Soham Bhattacharjee$^1$, Asif Ekbal$^{1, 2}$\\
  $^1$Department of Computer Science and Engineering, IIT Patna\\ $^2$School of AI and Data Science, IIT Jodhpur \\
  \texttt{dibyanayan\_2321cs14@iitp.ac.in, sohambhattacharjeenghss@gmail.com, asif@iitj.ac.in}
  \\}

\begin{document}
\maketitle
\begin{abstract}
Large Language Models (LLMs) are highly proficient in language-based tasks. Their language capabilities have positioned them at the forefront of the future AGI (Artificial General Intelligence) race. However, on closer inspection, \citet{valmeekam2024llmscantplanlrms, zečević2023causalparrotslargelanguage, wu2024reasoningrecitingexploringcapabilities} highlight a significant gap between their language proficiency and reasoning abilities. Reasoning in LLMs and Vision Language Models (VLMs) aims to bridge this gap by enabling these models to think and re-evaluate their actions and responses. Reasoning is an essential capability for complex problem-solving and a necessary step toward establishing trust in Artificial Intelligence (AI). This will make AI suitable for deployment in sensitive domains, such as healthcare,  banking, law, defense, security etc. In recent times, with the advent of powerful reasoning models like OpenAI O1 and DeepSeek R1, reasoning endowment has become a critical research topic in LLMs. In this paper, we provide a detailed overview and comparison of existing reasoning techniques and present a systematic survey of reasoning-imbued language models. We also study current challenges and present our findings.
\end{abstract}

\begin{center}
\textit{"Thinking is the hardest work there is, which is probably why so few engage in it."} \\
\hfill \textasciitilde Henry Ford 
\end{center}

\section{Introduction}

According to epistemology—the philosophy of knowledge—“reasoning” is the ability to draw inferences from evidence or premises \citep{sep-epistemology}. This capacity enables humans to think, acquire knowledge, and make informed decisions. While reasoning comes naturally to humans, emulating genuine reasoning in language models remains challenging. GPT-3 \citep{brown2020languagemodelsfewshotlearners} first exhibited low-level reasoning and instruction following, which was later advanced in ChatGPT (GPT-3.5) and GPT-4, achieving state-of-the-art performance across many tasks. The introduction of chain-of-thought prompting \cite{wei2023chainofthoughtpromptingelicitsreasoning}—which decomposes complex problems into manageable steps—further enhanced these models, prompting claims that GPT-4 shows “sparks of AGI” \cite{bubeck2023sparksartificialgeneralintelligence}.

However, research has shown that although LLMs can emulate reasoning on structured benchmarks, they do not truly reason as humans do \citep{valmeekam2024llmscantplanlrms, wei2023chainofthoughtpromptingelicitsreasoning}. For example, solving a complex mathematical proof requires decomposing the problem into subproblems and iteratively refining a solution—an ability that LLMs often struggle with without additional guidance. This insight has spurred efforts to evoke genuine reasoning in LLMs using techniques beyond simple scaling. Notably, OpenAI’s o1 models—despite having the same parameter count as GPT-3.5—outperform GPT-4 (10 times larger) on mathematical and reasoning tasks, indicating that scaling pre-training has diminishing returns. Consequently, researchers are turning to innovative test-time techniques that optimize pre-trained models.

Classical methods. such as Monte Carlo Tree Search (MCTS) and reward modeling, as used in AlphaGo \citep{Silver_2016}, are now re-imagined for LLMs. For instance, Reflexion \citep{shinn2023reflexionlanguageagentsverbal} allows an LLM to generate multiple chains of thought, evaluates them with a reward model, and iteratively refines its reasoning. This approach not only searches the model’s internal thought space for better conclusions but also enables self-training techniques that use generated reasoning traces as extra training data. These advances mark a shift from brute-force scaling toward leveraging innovative inference-time strategies for superior reasoning.

In this paper, we present a comprehensive review of current strategies for logical and critical thinking in LLMs. We subdivide the field with respect three key paradigms—reinforcement learning, test time computation, and self-training—applied during both training and inference. Our work stands out because:
\begin{enumerate}
    \item Unlike previous surveys like \cite{qiao2023reasoninglanguagemodelprompting, huang2023reasoninglargelanguagemodels}, our work is up-to-date with the latest advancements in LLM reasoning.
    \item It improves upon the existing surveys like \cite{kumar2025llmposttrainingdeepdive, li202512surveyreasoning} by providing a clear taxonomy that spans the entire field of reasoning. 
    \item Furthermore, our work is beginner-friendly and offers a top-down view of the key ideas alongside visual representations and detailed discussion of specific papers.
\end{enumerate}
This survey not only organizes diverse approaches into one coherent framework but also explains why advanced LLM reasoning is essential for complex problem solving, as demonstrated by tasks like intricate mathematical proofs.

\section{Preliminary Concepts}

This section provides an overview of the fundamental concepts underlying our work, including language model (LM) objectives, reinforcement learning (RL) principles, and Monte Carlo Tree Search (MCTS).

\subsection{Language Model Objectives and Sampling}

Language models generate text by predicting the next token given a prompt. Formally, given a prompt $x = (x_1, x_2, \dots, x_n)$ (where $x_i$ is a token at $i$th index,
a causal language model estimates the probability of the next token \(x_{n+1}\) as
\[
P_{\theta}(x_{n+1} \mid x_1, x_2, \dots, x_n).
\]
,where $\theta$ is the parameter of the language model. These models are trained using the \textit{causal language modeling }loss:
\[
\mathcal{L}_{\text{LM}}(\theta) = -\sum_{t=1}^{T} \log P_{\theta}(x_t \mid x_1, \dots, x_{t-1}),
\]
where \(T\) is the total number of tokens in a sequence. During inference, given a prompt \(x\), the model generates new tokens by sampling 

$$x_{n+1} \sim P_{\theta}(\cdot \mid x_1, \dots, x_n)$$

\subsection{Basics of Reinforcement Learning for LLMs}

Reinforcement learning (RL) trains an agent to interact with an environment to maximize a cumulative reward. In our framework, the agent is the LLM and the environment can be an external tool, a trained model, or the same LLM. Sometimes the environment is replaced with a \textit{world model}, which is model's internal representation of the environment that enables the agent to predict future states and rewards.

At time step \(t\), the agent observes a state \(s_t\) as a collection of previously generated tokens $x_1, \dots, x_{t-1}$ or reasoning steps, takes an action, which can be the next token or the next reasoning step $x_t$, and the environment in turn returns a reward \(r_t\). This agent-environment interaction is depicted in the block diagram below (Figure \ref{fig:basic-rl}):

\begin{figure}[h]
    \centering
    \includegraphics[width=\linewidth]{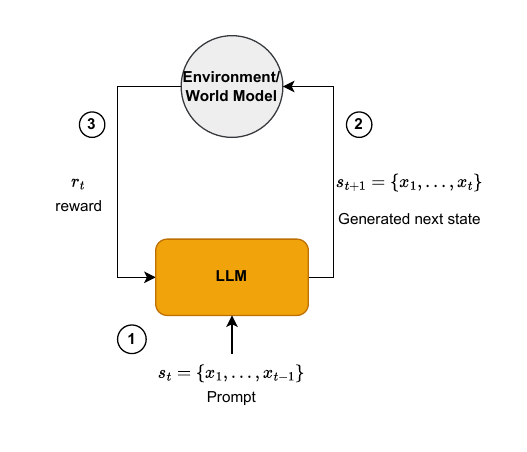}
    \caption{Basic Reinforcement learning setup consisting of an environment and an LLM.}
    \label{fig:basic-rl}
\end{figure}


\subsubsection{Policy Optimization}

The LLM token output distribution is called a policy $\pi_{\theta}(a \mid s)$ (parameterized by $\theta$), where the objective of RL is to learn a policy such that the following expected cumulative reward is maximized.
\[
J(\pi_{_{\theta}}) = \mathbb{E}_{\pi_\theta}\left[\sum_{t=0}^{\infty} \gamma^t r_t\right],
\]
where \(\gamma \in [0,1)\) is the discount factor.

A simple approach to optimize the policy is the vanilla policy gradient:
\[
\nabla_\theta J(\pi_\theta) = \mathbb{E}_{\pi_\theta}\left[\nabla_\theta \log \pi_\theta(a_t \mid s_t) \, G_t\right],
\]
with the return \(G_t = \sum_{t'=t}^{\infty} \gamma^{t'-t} r_{t'}\). More advanced methods such as Proximal Policy Optimization (PPO) \cite{schulman2017proximalpolicyoptimizationalgorithms} and GRPO \cite{shao2024deepseekmathpushinglimitsmathematical} build upon this by incorporating constraints to ensure stable and efficient updates.
The goal of RL is to obtain the \textit{optimal policy $\pi_{\theta^*}$} that obtains the highest return by iteratively optimizing $J(\pi_\theta)$.

\textbf{Value Function.} Value functions are of two kinds, i) \textit{State value function} ($V(s)$) which measures the expected returns if we start from state $s$ and act according to the optimal policy, ii) \textit{Action value function} ($Q(s,a)$) which measures the expected returns if we start from state $s$ and take action $a$ (can be any action) and then follow the optimal policy. In summary, value function is a function of a state (thus can be parameterized by a neural network) which outputs the importance of that state. The rule of thumb is to select the state with the highest value.

\subsection{Preference Optimization}

Preference Optimization (PO) directly aligns model behavior with preference data. Given an input \(x\) and two candidate outputs \(y^+\) and \(y^-\) (where \(y^+\) is preferred over \(y^-\)), DPO \cite{rafailov2024directpreferenceoptimizationlanguage} adjusts the model so that the probability of generating \(y^+\) exceeds that of \(y^-\). The objective (based on Bradley-Terry model) is defined as:
\[
\mathcal{L}_{\text{PO}}(\theta) = - \sum_{(x, y^+, y^-)} \frac{e^{r(y^+, x)}}{e^{r(y^+, x)} + e^{r(y^-, x)}}
\]

where $r(y, x)$ denotes the reward assigned  to output \(y\) with \(x\) as the prompt. This loss encourages the model to assign higher likelihood to preferred outputs, effectively refining its reasoning process based on observed preferences.

\subsection{Monte Carlo Tree Search (MCTS)}

MCTS is a search algorithm that uses random sampling to explore a decision tree and guide action selection. The key steps of MCTS are:
\begin{enumerate}
    \item \textbf{Selection:} Starting at the root node, recursively select child nodes based on a policy (e.g., Upper Confidence Bound for Trees (UCT), or maximum action-value $Q$ or a combination thereof) until reaching a leaf node.  
    \item \textbf{Expansion:} If the leaf node is non-terminal, expand it by adding one or more child nodes. The associated position of the newly expanded node is evaluated using a value network. 
    \item \textbf{Simulation:} Perform a rollout from the new node using random or policy-guided sampling to estimate the outcome.
    \item \textbf{Backpropagation:} Update the statistics (e.g., visit count, average reward or $Q$ values) of nodes along the path based on the simulation result.
\end{enumerate}
MCTS efficiently explores the search space by concentrating computational resources on the most promising branches.

\section{Methods}

The overall method landscape of achieving reasoning in large language models (LLMs) is broken down into three overall methods: i) Reinforcement Learning, ii) Test Time Compute and iii) Self-Training Methods. Each of them is described in detail below.

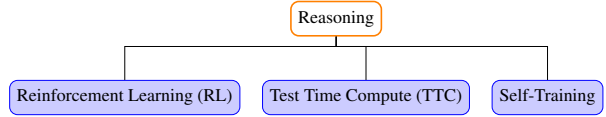
\begin{figure}[h]
\centering
\resizebox{0.5\textwidth}{!}{
\begin{forest}
for tree={
    draw,
    rounded corners,
    align=center,
    child anchor=north,
    grow =south,
    parent anchor=south,
    font = \small,
    l sep=2em,
    s sep=1em,
    edge path={%
      \noexpand\path[\forestoption{edge}] 
        (!u.parent anchor) -- +(0,-5pt) -| (.child anchor)
        \forestoption{edge label};},%
    if n children=0 {fill=blue!20, draw=blue}{fill=white, draw=orange, thick}
}
[Reasoning
    [Reinforcement Learning (RL)]
    [Test Time Compute (TTC)]
    [Self-Training]
]
\end{forest}
}
\caption{Taxonomy of methods for reasoning.}
\end{figure}

\subsection{Reinforcement Learning}

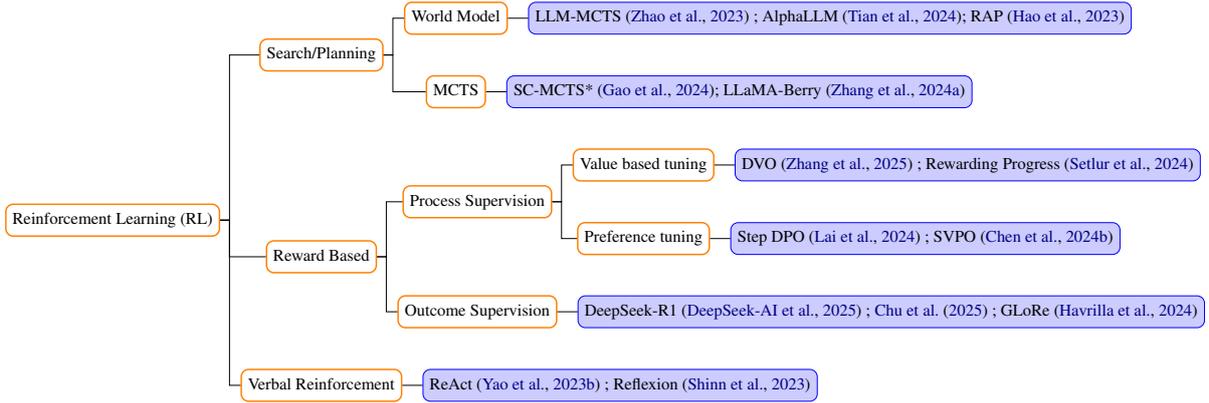
\begin{figure*}[ht]
\centering
\resizebox{\textwidth}{!}{
\begin{forest}
for tree={
    draw,
    rounded corners,
    align=left,
    child anchor=west,
    grow =east,
    parent anchor=east,
    font = \small,
    l sep=1em,
    s sep=2em,    
    edge path = {\noexpand\path[\forestoption{edge}](!u.parent anchor) -- +(5pt,0) |- (.child anchor)\forestoption{edge label};},    
    if n children=0 {fill=blue!20, draw=blue}{fill=white, draw=orange, thick}
}
    [Reinforcement Learning (RL)
        [Verbal Reinforcement 
            [ReAct \cite{yao2023reactsynergizingreasoningacting} ; Reflexion \cite{shinn2023reflexionlanguageagentsverbal}]           
        ]
        [Reward Based
            [Outcome Supervision
                [DeepSeek-R1 \cite{deepseekai2025deepseekr1incentivizingreasoningcapability} ; \citet{chu2025sftmemorizesrlgeneralizes} ; GLoRe \cite{havrilla2024glorewhenwhereimprove}]
            ]
            [Process Supervision
                [Preference tuning
                    [Step DPO \cite{lai2024stepdpostepwisepreferenceoptimization} ; SVPO \cite{chen2024steplevelvaluepreferenceoptimization}]
                ]
                [Value based tuning
                    [DVO \cite{zhang2025directvalueoptimizationimproving} ; Rewarding Progress \cite{setlur2024rewardingprogressscalingautomated}]
                ]
            ]
        ]
        [Search/Planning 
            [MCTS
                [SC-MCTS* \cite{gao2024interpretablecontrastivemontecarlo}; LLaMA-Berry \cite{zhang2024llamaberrypairwiseoptimizationo1like}]
            ]
            [World Model
                [LLM-MCTS \cite{zhao2023largelanguagemodelscommonsense} ; AlphaLLM \cite{tian2024selfimprovementllmsimaginationsearching}; RAP \cite{hao2023reasoninglanguagemodelplanning}]
            ]
        ]
    ]
\end{forest}
}
\caption{Taxonomy of Reasoning with Reinforcement Learning.}
\end{figure*}

 In any reasoning-based system where the objective is to reach a goal from a stated starting point, the optimal policy tends to select the path with the highest expected reward. This path consists of intermediate steps—commonly referred to as reasoning steps. For instance, in AlphaGo \cite{Silver_2016} the optimal next moves are considered as reasoning steps, whereas in solving a math problem, the intermediate steps constitute the reasoning process. Consequently, reinforcement learning (RL) can be a powerful strategy for eliciting reasoning in language models. 

These strategies can be broadly divided into three categories: i) \textbf{Verbal Reinforcement},  ii) \textbf{Reward-based Reinforcement}, which can be further subdivided into \textit{process supervision} and \textit{outcome supervision}, and iii) \textbf{Search/Planning}.

\begin{figure}[h]
    \centering
    \includegraphics[width=\linewidth]{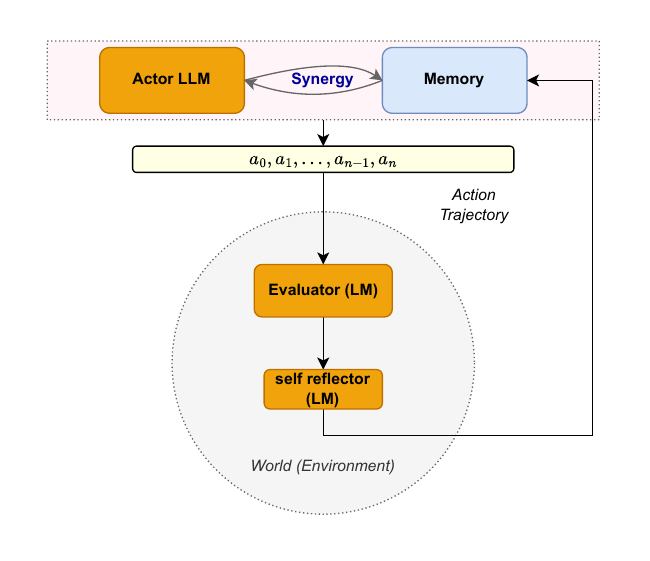}
    \caption{Verbal reinforcement objective that relies on generation followed by NL feedback. Often the feedback consists of just observing the impact of actor LLM's own action to the environment, as in \cite{yao2023reactsynergizingreasoningacting}}
    \label{fig:vrl}
\end{figure}

In recent years, hybrid approaches that combine two or more of these components have also been explored. For example, verbal RL can be combined with tree search as in RAP \cite{hao2023reasoninglanguagemodelplanning}, or process supervision can be combined with Monte Carlo tree search as demonstrated in rStar-Math \cite{guan2025rstarmathsmallllmsmaster}.

i) \textbf{Verbal Reinforcement:} In verbal reinforcement, a base language model (LM) connected to a memory (either in-context or external) produces an action trajectory given a prompt. This action (i.e., a sequence of natural language tokens) is fed to an environment, where two LM-based modules—referred to as the \emph{Evaluator} and the \emph{Self-Reflector}—provide feedback in natural language on the generated reasoning trace. This feedback is stored in the memory and subsequently used to guide further generation. An overall diagram of this approach is shown in Figure~\ref{fig:vrl}. For instance, \cite{yao2023reactsynergizingreasoningacting} introduced a strategy that employs a thought-action-observation triplet to enhance reasoning. Similarly, the Reflexion framework \cite{shinn2023reflexionlanguageagentsverbal} explores the use of verbal feedback in language agents.

\begin{figure}[ht]
\centering
\includegraphics[width=\linewidth]{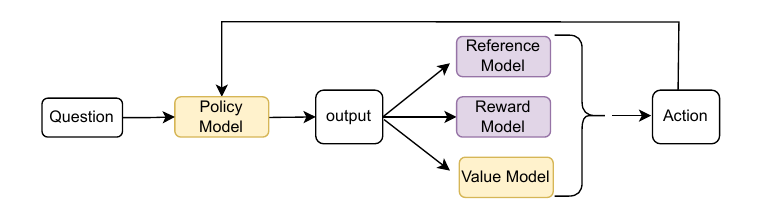}
\includegraphics[width=\linewidth]{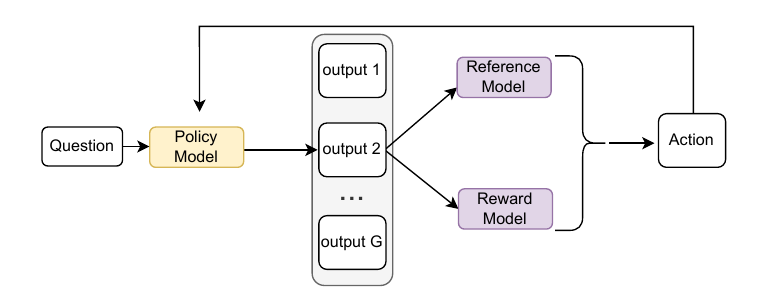}

\captionsetup{justification=raggedright,singlelinecheck=false}

\caption{Improvement of GRPO over PPO. \textbf{Up:} DialCot \cite{han-etal-2023-dialcot} uses PPO for policy updation. 
\textbf{Down:} In \citet{shao2024deepseekmathpushinglimitsmathematical} GRPO foregoes the value model 
and calculates the final reward from group scores of multiple outputs, significantly reducing training resources}
\label{fig:grpo-ppo}
\end{figure}

\begin{figure*}[htbp]
  \centering
  \begin{subfigure}[b]{0.32\textwidth}
    \centering
    \includegraphics[width=\linewidth]{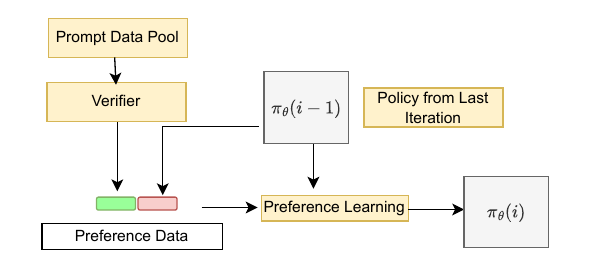}
    \caption{\textbf{DPO} \cite{rafailov2024directpreferenceoptimizationlanguage} directly optimizes the policy based on instance level preferences. As a result it misses out on granular step level preferences}
    \label{fig:1}
  \end{subfigure}
  \begin{subfigure}[b]{0.32\textwidth}
    \centering
    \includegraphics[width=\linewidth]{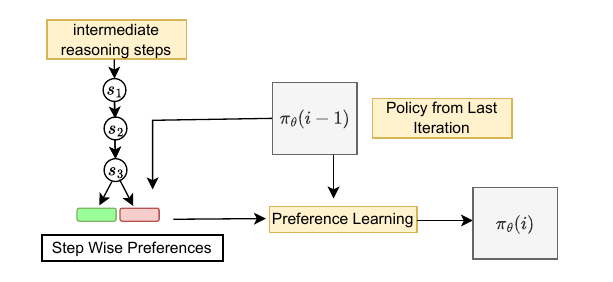}
    \caption{\textbf{Step-DPO} \cite{lai2024stepdpostepwisepreferenceoptimization} uses step level reasoning preferences to improve upon DPO}
    \label{fig:2}
  \end{subfigure}
  \begin{subfigure}[b]{0.32\textwidth}
    \centering
    \includegraphics[width=\linewidth]{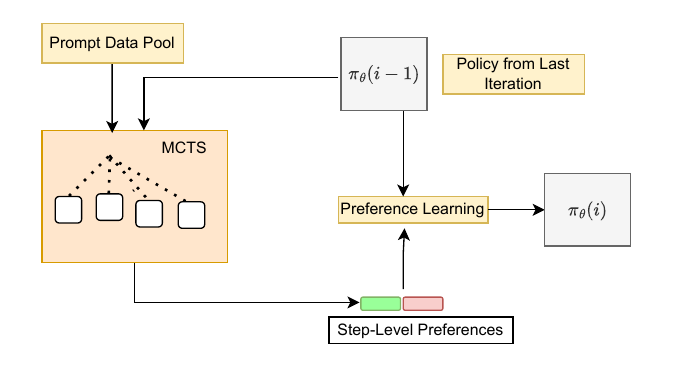}
    \caption{\textbf{MCTS} combined with \textbf{DPO} \cite{xie2024montecarlotreesearch} also uses step level preferences from actions estimated by MCTS to assign the preferences}
    \label{fig:3}
  \end{subfigure}
  \caption{DPO and its variants which use step-wise preference data for granular reasoning tasks.}
  \label{fig:three_images}
\end{figure*}

ii) \textbf{Reward based Reinforcement:} These methods can be broadly divided into two categories:

\begin{enumerate}
    \item \textit{Process Supervision:} In this approach, each reasoning step (note that the reasoning step can either be the next token or the next step) is individually rewarded based on its relevance to forming an optimal reasoning chain. Since, the intermediate steps are rewarded instead of only the final outcome, the model moves towards a more coherent final answer.



    In DialCoT \cite{han-etal-2023-dialcot}, PPO was first employed to invoke chain-of-thought reasoning in language models. Although PPO typically relies on \textit{outcome supervision}—because it updates the policy only after seeing a final outcome, DialCoT divides each question into multiple smaller dialogue‐based sub‐questions that each receive rewards, effectively transforming PPO into a \textit{process‐supervised} method. Building on this idea, VinePPO \cite{kazemnejad2024vineppounlockingrlpotential} replaces the value‐based network in PPO with Monte Carlo simulations of intermediate steps, making the algorithm more computationally  efficient and less error‐prone. In \citet{chen2024alphamathzeroprocesssupervision}, the authors demonstrate that process supervision can be achieved without explicit preference data by training a value network to predict the value of intermediate states using Monte Carlo estimates of the final reward. Likewise, \cite{luo2024improvemathematicalreasoninglanguage} also combines process supervision with MCTS for search. 

    Acquiring process-supervised data is often challenging, and training a value-based network is computationally expensive. To address this, researchers introduced the concept of Direct Preference Optimization (DPO) in \citet{rafailov2024directpreferenceoptimizationlanguage}. DPO employs an \textit{outcome-supervised} strategy, where positive and negative preference data pairs are directly used to update the model's policy without relying on an external value or reward model. While DPO gained popularity for its simplicity and effectiveness in chat benchmarks, it struggled with reasoning-based tasks due to the lack of granular process supervision, as noted in \citet{lai2024stepdpostepwisepreferenceoptimization}.
    Researchers have successfully tackled this problem by combining process supervision with preference optimization in \citet{lai2024stepdpostepwisepreferenceoptimization}, where step-level preference data is used for policy updation. The quality of these self-generated pairs can be enhanced by performing MCTS rollouts from the root node \( s_0 \) and constructing intermediate-state preference pairs \((s_i \succ s_j)\), where \( s_i \) is preferred over \( s_j \) if it leads to the correct answer more frequently. Using MCTS also gets rid of the expensive human or language model supervision in annotating preference pairs. These ideas are further explored in \citet{jiao2024learningplanningbasedreasoningtrajectories, xie2024montecarlotreesearch, chen2024steplevelvaluepreferenceoptimization}, with an excellent summary provided in \citet{li2025enhancingreasoningprocesssupervision}. Due to the labor-intensive nature of human supervision for constructing preference data, Direct Value Optimization (DVO) \cite{zhang2025directvalueoptimizationimproving} has been introduced. Much of the current research focuses on enhancing process rewards through better credit assignment strategies \cite{setlur2024rewardingprogressscalingautomated, cui2025processreinforcementimplicitrewards, zhang2024entropyregularizedprocessrewardmodel}. Figure \ref{fig:three_images} illustrate classic DPO and its improvements for handling step level granular preference pairs which aid in better reasoning.

    
    
    \item \textit{Outcome Supervision:} While \cite{lightman2023letsverifystepstep} originally showed that process supervision outperforms outcome supervision, this view is challenged by superior performance of models such as Deepseek-R1 \cite{deepseekai2025deepseekr1incentivizingreasoningcapability}, which rely solely on outcome-based rewards. This is also supported by the GRPO (Group Relative Policy Optimization) algorithm \cite{shao2024deepseekmathpushinglimitsmathematical}, which improves upon PPO by replacing the expensive value-based network with group relative scores during policy updates. See Figure \ref{fig:grpo-ppo} for a comparison between GRPO and PPO. In \citet{chu2025sftmemorizesrlgeneralizes}, the authors demonstrate that outcome-based rewards can even outperform supervised fine-tuning (SFT) for reasoning tasks. For disciplines such as mathematics, science, or coding---where refined reasoning is essential---traditional outcome reward models (ORM) or preference reward models (PRM) suffer from either sparse reward signals or the need for extensive human annotation. To address these issues, \citet{havrilla2024glorewhenwhereimprove} presents a method that trains a step-wise ORM (s-ORM) using synthetic data. This approach not only alleviates the data-hungry nature of PRMs but also more accurately detects incorrect reasoning steps, thereby mitigating the sparse reward and credit assignment challenges inherent in standard ORMs.
\end{enumerate}

\begin{figure}
    \centering
    \includegraphics[width=\linewidth]{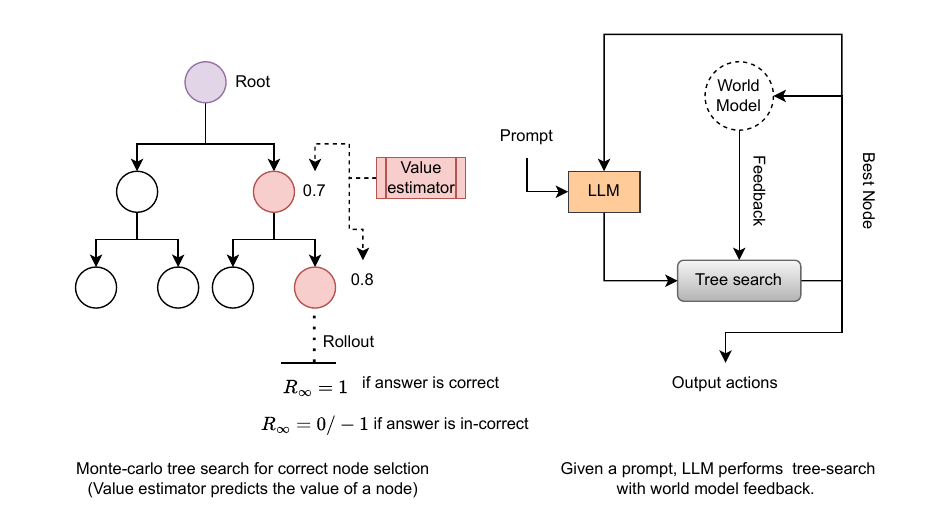}
    \caption{\textbf{Left:} Monte Carlo tree search where the state values are evaluated using a value estimator network which is trained using RL via preference data or monte-carlo estimate of the reward. \textbf{Right:} World model as a feedback mechanism for tree-search.}
    \label{fig:mcts}
\end{figure}

iii) \textbf{Search:} Search-based techniques in reinforcement learning leverage tree search methods—most notably Monte Carlo Tree Search (MCTS) and its variants—to improve decision-making. This approach was pioneered in systems like AlphaGo and its subsequent variants, where MCTS is integrated with a value and policy network (the LLM) to explore potential moves efficiently \cite{gao2024interpretablecontrastivemontecarlo}. The same underlying idea—using search to simulate future outcomes and guide action selection—has been extended to domains requiring verifiable results, such as mathematics.
The value network that estimates the value of potential moves in tree search and can be trained using: i) preference data collected from MCTS rollout. Concretely, $s_i$ is considered preferred than $s_j$ (denoted by $s_i \succ s_j$) if $s_i$ leads to the correct answer more often than $s_j$. \cite{zhang2024llamaberrypairwiseoptimizationo1like, ma2025steplevelrewardmodelsrewarding}, ii) Using Monte-Carlo estimate of the overall reward achieved from the current state \cite{chen2024alphamathzeroprocesssupervision}.


More recently, generic problem solving has benefited from frameworks that treat large language models (LLMs) as world models. In these setups, an LLM generates predictions about the environment, and this world model is augmented with MCTS or similar search strategies to systematically explore potential solutions \cite{ zhao2023largelanguagemodelscommonsense, feng2024alphazeroliketreesearchguidelarge, tian2024selfimprovementllmsimaginationsearching, zhou2024languageagenttreesearch}. Figure \ref{fig:mcts} explains the overall details.

\subsection{Test Time Compute}

\begin{figure*}[t]
\centering
\resizebox{\textwidth}{!}{
\begin{forest}
for tree={
    draw,
    rounded corners,
    align=left,
    child anchor=west,
    grow =east,
    parent anchor=east,
    l sep=1em,
    s sep=2em,
    edge path = {\noexpand\path[\forestoption{edge}](!u.parent anchor) -- +(5pt,0) |- (.child anchor)\forestoption{edge label};},
    if n children=0 {fill=blue!20, draw=blue}{fill=white, draw=orange, thick}
}
[Test Time Compute (TTC)
    [Feedback Guided Improvement
        [Generation time feedback
            [Step Feedback
                [GRACE \cite{khalifa2023gracediscriminatorguidedchainofthoughtreasoning} ; CoRE \cite{zhu-etal-2023-solving}]
            ]
            [Outcome Feedback
                [LEVER \cite{ni2023leverlearningverifylanguagetocode} ; CodeT \cite{chen2022codetcodegenerationgenerated}]                
            ]
        ]
        [Post-hoc Feedback
            [LLM based Debate 
                [Debate\cite{du2023improvingfactualityreasoninglanguage}; CMD \cite{wang2024rethinkingboundsllmreasoning}; LMvLM \cite{cohen2023lmvslmdetecting}; ReConcile \cite{chen2024reconcileroundtableconferenceimproves}]
            ]
            [Trained Model Feedback   
                [DeCRIM \cite{ferraz2024llmselfcorrectiondecrimdecompose} ; Re3 \cite{yang2022re3generatinglongerstories}]
            ]
        ]
    ]
    [Scaling Adaptation
        [Compute Scaling
            [Chain-of-Thought \cite{wei2023chainofthoughtpromptingelicitsreasoning} ; Forest-of-Thought \cite{bi2025forestofthoughtscalingtesttimecompute}; Graph-of-Thought \cite{Besta_2024} ; Buffer-of-Thought \cite{yang2024bufferthoughtsthoughtaugmentedreasoning}]
        ]
        [Self Feedback
            [Natural Language
                [Self-Refine \cite{madaan2023selfrefineiterativerefinementselffeedback} ; \citet{weng2023largelanguagemodelsbetter}]
            ]
            [Score based search
                [Deductive Beam Search \citet{zhu2024deductivebeamsearchdecoding} ; AlphaLLM \cite{tian2024selfimprovementllmsimaginationsearching}]
            ]
        ]        
    ]         
]    
\end{forest}
}
\caption{Taxonomy of Reasoning with Test Time Compute (TTC)}
\end{figure*}
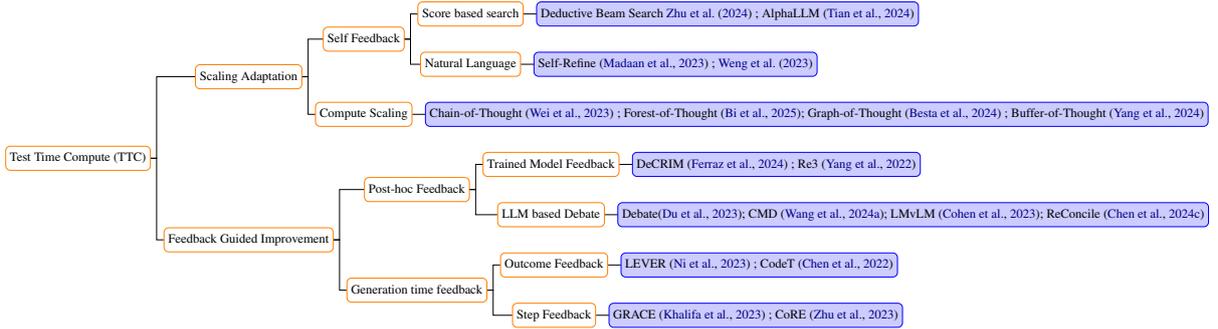

Modern LLMs have demonstrated impressive reasoning abilities; however, their fixed parameters and static inference can limit their capacity to adapt to new or complex tasks on the fly. Test time compute is motivated by the need to overcome these limitations by allowing models to refine or extend their reasoning capabilities during deployment. \textit{In test time compute setting, the pretrained LLM is never trained by either supervised or reinforcement learning-based techniques.}

Test-time capabilities can be broadly categorized into: 
\begin{enumerate}
    \item \textbf{Feedback Guided Improvement:} In this category, the generation from the model is refined based on external feedback provided by various critique modules (e.g., program execution modules, external knowledge bases, or trained models).
    \item \textbf{Scaling/Adaptation:} In this approach, the compute allocated at test time is scaled up to enhance the quality of the output.
\end{enumerate}

\subsubsection{Feedback Guided Improvement}

\begin{figure}
    \centering
    \includegraphics[width=\linewidth]{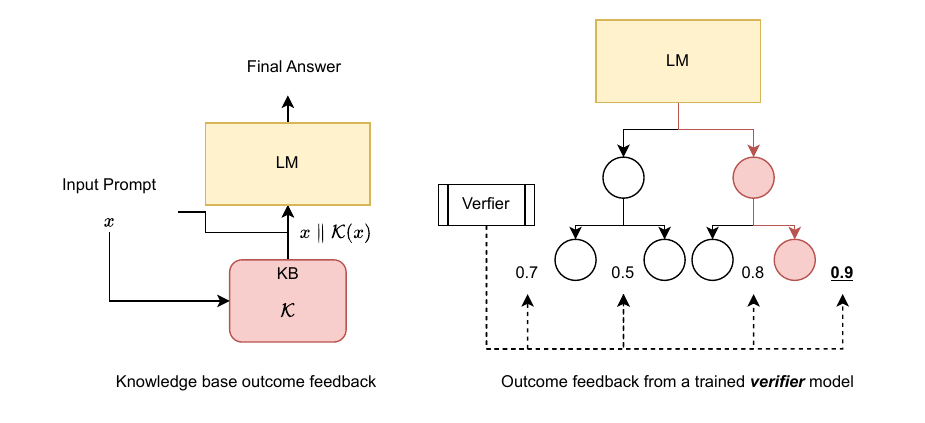}
    \caption{\textbf{Left:} The input prompt is passed to a knowledge base that provides feedback, which is then fed back into the LM for further output. \textbf{Right:} Multiple branches are explored from a start node (generated by the LM), and only the branch with the highest estimated verifier feedback score is selected.}
    \label{fig:of_survey}
\end{figure}

Depending on the timing of the feedback, this paradigm can be further grouped into two subdomains: 
\begin{enumerate}
    \item \textbf{Generation Time Feedback:} In this case, the LLM receives feedback during the generation process, which can be in the form of numerical scores assessing either partial or complete outputs.
    \begin{enumerate}
        \item \textbf{Step-Feedback (SF):} Here, beam search or Monte Carlo Tree Search (MCTS) is employed in which a critique module provides feedback at each step of the reasoning process. Only the highest scoring reasoning chains are retained. Figure~\ref{fig:sf_survey} illustrates how an external verifier scores intermediate states in both beam search and MCTS.
        \item \textbf{Outcome-Feedback (OF):} In this approach, multiple generations are produced and then scored by an external critique module. Only the generations with the highest scores are retained. Figure~\ref{fig:of_survey} depicts a feedback mechanism where the input prompt is first enriched with feedback from an external knowledge base or verifier, and then multiple branches are explored with the highest scoring branch being selected.
    \end{enumerate}
    
    Depending on the application, multiple feedback-based models have been proposed.
    
    \medskip
    \noindent
    \textit{Code Execution as Critique:} For example, CodeT \cite{chen2022codetcodegenerationgenerated} employs the same language model to generate multiple code candidates along with test cases, selecting the final code only if all test cases are passed. Similarly, LEVER \cite{ni2023leverlearningverifylanguagetocode} uses a program executor to verify the correctness of the generated code and provide feedback. Both methods operate as outcome-feedback based critiques. For using various kinds of tools as feedback, \cite{gou2024criticlargelanguagemodels} proposes interactive tool use as a critiquing method. Similarly, \citet{chen2023teachinglargelanguagemodels} demonstrates using results of generated code and a few in-context example as feedback, large language model learns to debug its predicted program.
    
    \medskip
    \noindent
    \textit{External Knowledge as Critique:} In this setup, an external knowledge base $\mathcal{K}$ is consulted before generating an answer. The feedback from $\mathcal{K}$, denoted as $\mathcal{K}(x)$, may be provided as a natural language sentence or as discrete scores. For example, MemPrompt \cite{madaan-etal-2022-memory} utilizes a pool of prior user feedback (serving as a knowledge base) to guide text generation based on the current prompt. In another example, Varshney et al. leverage self-inquiry and web-search to retrieve relevant information to correct hallucinated statements, where the knowledge base comprises both the language model and external web data.
    
    \medskip
    \noindent
    \textit{Trained Model as Critique:} A trained model can be used as a critique mechanism in either a step-feedback or outcome-feedback configuration.
    
    \begin{figure*}
        \centering
        \includegraphics[width=\textwidth]{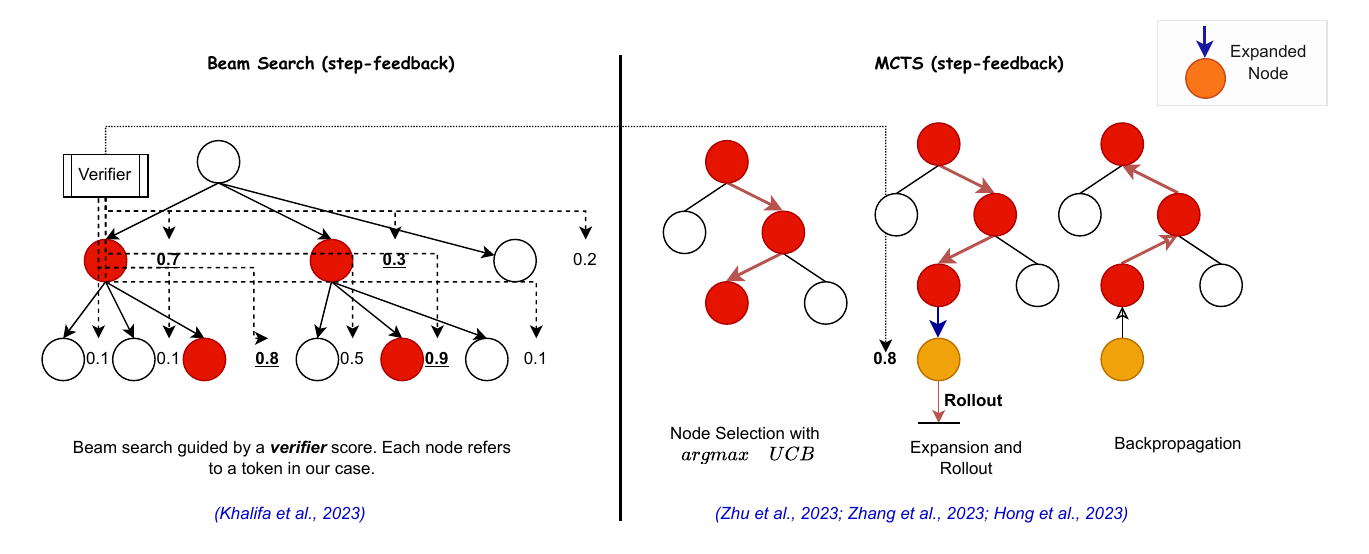}
       \caption{Step-feedback illustration using Beam search and MCTS. In Beam search, an external verifier scores intermediate nodes to guide selection. In MCTS, a node is expanded and a rollout begins from that node; here, the verifier assigns it a value of 0.8.}

        \label{fig:sf_survey}
    \end{figure*}
    
    GRACE \cite{khalifa2023gracediscriminatorguidedchainofthoughtreasoning} employs beam search with step-feedback, guiding the chain-of-thought (CoT) process using a discriminator. CoRE \cite{zhu-etal-2023-solving} uses MCTS for math problem solving, where the selection operation at each node is informed by a predicted reward from a trained model critique. DIVERSE \cite{li-etal-2023-making} uses binary-valued outcome-feedback based on a trained DeBERTa model, significantly improving the problem-solving rate on GSM8K from 17.9\% to 58.1\%. Additionally, \cite{zhang2023planninglargelanguagemodels} proposes a transformer decoding mechanism based on MCTS for code generation, with reward signals obtained from executing generated code on public test cases. Similarly, \cite{hong2023faithfulquestionansweringmontecarlo} employs MCTS for common-sense reasoning by constructing an entailment tree.
    
    \item \textbf{Post-hoc Feedback:} After the base LLM generates its outputs, these outputs can be refined in a post-hoc manner using separate models.
    \begin{enumerate}
        \item \textit{LLM-Based Debate:} Two LLMs—framed as an examinee and an examiner—engage in a multi-turn debate where the original answer is discussed and refined. Originally proposed in \cite{du2023improvingfactualityreasoninglanguage}, this approach has been adopted in several subsequent works \cite{cohen2023lmvslmdetecting, chen2024reconcileroundtableconferenceimproves}. Although promising, this approach has also faced criticism \cite{wang2024rethinkingboundsllmreasoning}.
        \item \textit{Trained Model Feedback and Refinement:} In this method, feedback in the form of either scalar values or natural language is used to revise the generated response in an inference-only setting \cite{yang2022re3generatinglongerstories}. Recently, generate-critique-refine pipelines have gained popularity \cite{ferraz2024llmselfcorrectiondecrimdecompose, wadhwa2024learningrefinefinegrainednatural}.
    \end{enumerate}
    
    Both of these post-hoc feedback mechanisms are described comprehensively in the survey of \cite{pan-etal-2024-automatically} (see Sections 3.2 and 3.3).
\end{enumerate}

\subsubsection{Scaling Test-Time Computation}

Recent works have shown that scaling test-time computation—such as best-of-N sampling—can outperform scaling during training. In this section, we focus on strategies that increase reasoning capabilities at test time by investing more computation, either by scaling up token-level compute, employing test-time training, or using self-feedback. Note that self-feedback differs from self-teaching: in self-feedback, the original model output is refined without further training of the base model. \textit{Scaling up token compute} and \textit{Self-feedback} are two mainstream strategies.

\textbf{\textit{Scaling Up Token Compute:}}  
Scaling up token-level compute involves generating multiple intermediate token outputs (i.e., reasoning steps) in parallel, allowing the exploration of various plausible reasoning pathways. This strategy is typically implemented via chain-of-thought (CoT) prompting and its subsequent variants. For instance, CoT \cite{wei2023chainofthoughtpromptingelicitsreasoning} generates reasoning in a linear, step-by-step manner, whereas tree-of-thought (ToT) \cite{yao2023treethoughtsdeliberateproblem} generalizes this by exploring multiple reasoning branches simultaneously and selecting the best one based on an LM-generated heuristic score. Due to the inherent noise in LM-generated scores, pairwise comparisons have been proposed \cite{zhang2024generatingchainofthoughtspairwisecomparisonapproach} to more reliably identify promising intermediate thoughts.

Further generalizations include forest-of-thought (FoT) \cite{bi2025forestofthoughtscalingtesttimecompute}, which employs sparse activation to select the most relevant reasoning paths among multiple trees, and graph-of-thought (GoT) \cite{Besta_2024}, which uses graph-based operations to aggregate and refine thoughts. Recently, buffer-of-thoughts \cite{yang2024bufferthoughtsthoughtaugmentedreasoning} has been introduced, where a meta-buffer stores high-level meta-thoughts that are dynamically instantiated during problem solving.

An alternative to pure scaling is integrating search with CoT. For example, CoT with search \cite{zhu2024deductivebeamsearchdecoding} uses a verifier at each reasoning step to check deductibility and mitigate error accumulation. In inductive reasoning problems, another approach constructs a set of natural language hypotheses that are then implemented as verifiable Python programs, as explored in hypothesis search \cite{wang2024hypothesissearchinductivereasoning}.

\begin{figure*}[t]
    \centering
    \includegraphics[width=\textwidth]{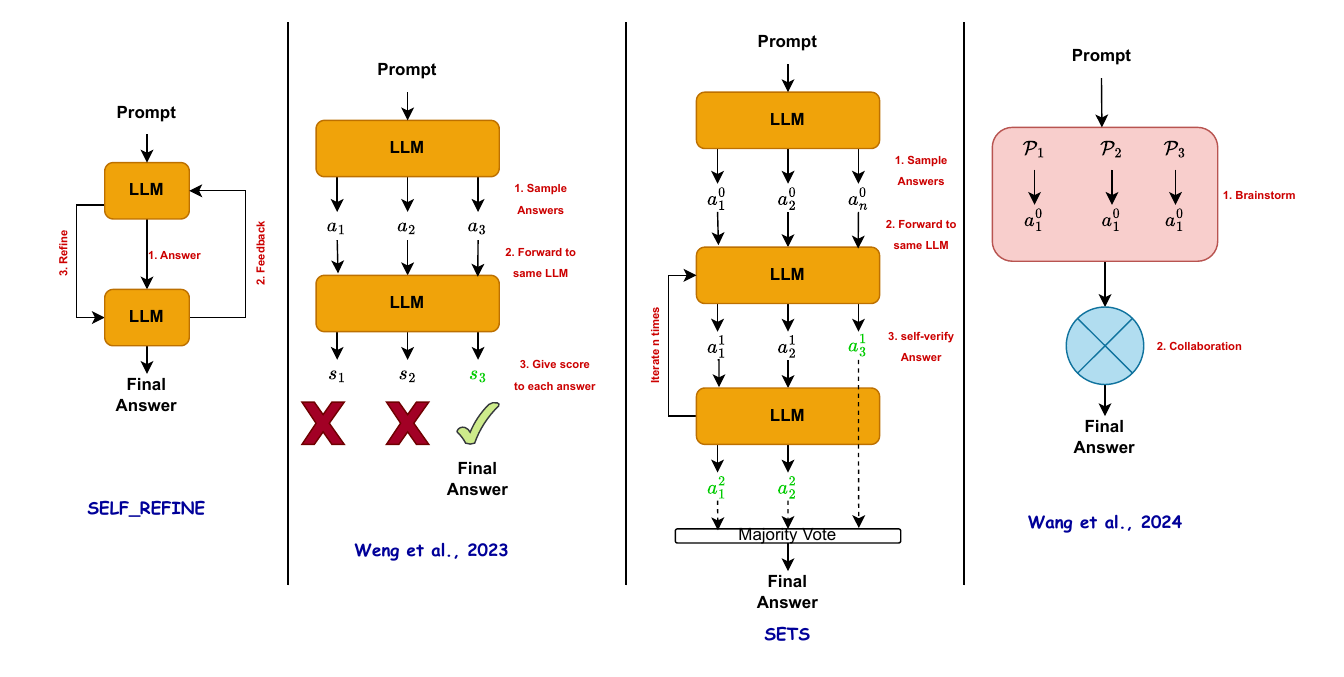}
    \caption{From left to right, we elaborate four main papers based on self-refinement objective. $a^i_j$ refers to the $i$th iteration and $j$ the sample answer. Green color denotes the particular answer is self-verified as correct.}
    \label{fig:self-refinement}
\end{figure*}

\textbf{\textit{Self-Feedback:}} Self-feedback can be utilized in two forms:
\begin{enumerate}
    \item \textbf{Natural Language Self-Feedback:} 
    Self-refine \cite{madaan2023selfrefineiterativerefinementselffeedback} is a pedagogical work that demonstrates how natural language feedback generated by the same LLM can be used to refine its reasoning process. This approach follows three steps: i) Answer Generation, ii) Feedback Generation, ii) Refinement.
    All steps are executed in a loop using the same LLM. As an extension, \citet{weng2023largelanguagemodelsbetter} proposed a scoring-based approach where multiple sampled answers are scored by the same LLM, and the answer with the highest score is selected as the final answer. In this method, separate conclusions are generated for each answer, followed by backward verification on those conclusions. Another strategy \cite{chen2025setsleveragingselfverificationselfcorrection} iteratively samples and applies self-verification until all sampled answers are confirmed as correct by the model; the final answer is then determined via majority voting among the verified responses. Additionally, self-collaboration has been introduced by repurposing a single LLM into a cognitive synergist with multiple personas \cite{wang2024unleashingemergentcognitivesynergy}. Here, multiple personas engage in a brainstorming session by generating separate answers, which are subsequently refined through inter-persona collaboration. The overall ideas of these approaches are summarized in Figure~\ref{fig:self-refinement}.
    
    \item \textbf{Self-Feedback as a Score for Tree-Search:} 
    Self-feedback can also be leveraged as a reward or scoring signal within search-based techniques. For instance, \citet{zhu2024deductivebeamsearchdecoding} propose a decoding algorithm that employs beam search for reasoning based on self-feedback. Other studies have demonstrated that MCTS combined with LLM-guided self-feedback can enhance the reasoning process \cite{tian2024selfimprovementllmsimaginationsearching, hao2023reasoninglanguagemodelplanning}. Furthermore, some methods reframe the LLM as a world model to steer an MCTS-based reasoning process \cite{zhao2023largelanguagemodelscommonsense}. The underlying technical framework in these self-feedback search techniques is analogous to that shown in Figure~\ref{fig:sf_survey}, with the distinction that the external step-feedback is replaced by self-generated feedback.
\end{enumerate}

ReVISE \cite{lee2025reviselearningrefinetesttime} exemplifies a hybrid approach by using a structured, curriculum-based preference learning method to both self-teach the base model and integrate test-time scaling via self-verification and correction. This approach is further enhanced by a confidence-aware decoding mechanism that leverages self-feedback. 

There are, however, criticisms suggesting that LLMs may not effectively aid in planning \cite{valmeekam2023planningabilitieslargelanguage} or self-correct their reasoning \cite{huang2024largelanguagemodelsselfcorrect}.

\subsection{Self-Training Methods}

\begin{figure*}
    \centering
    \includegraphics[width=\linewidth]{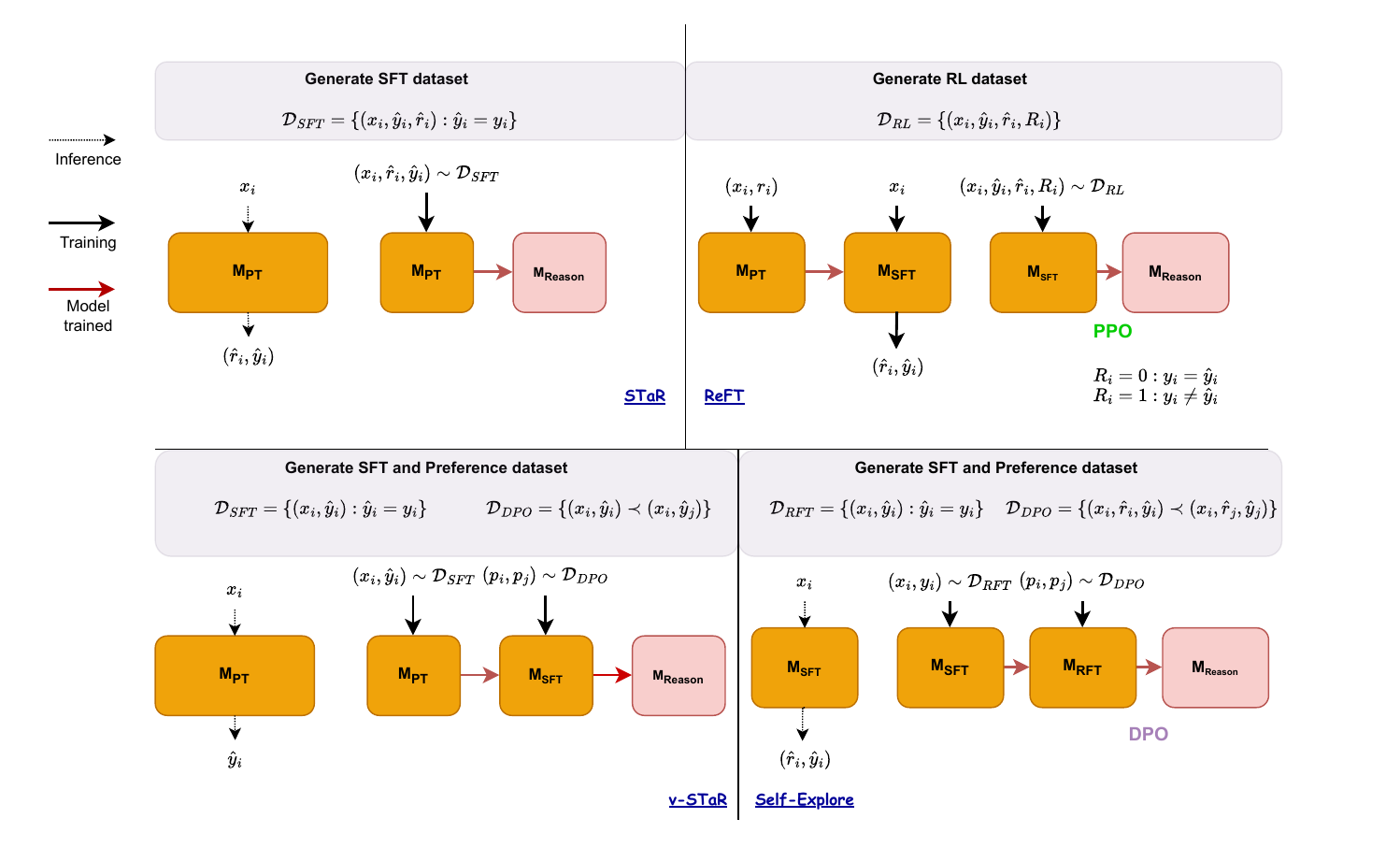}
    \caption{Four representative self-training methods (i.e. STaR, v-STaR, ReFT, Self-Explore) with training details.}
    \label{fig:self-taughts}
\end{figure*}

Pre-trained language models have revolutionized natural language processing; however, their fixed weights can become a bottleneck when faced with the dynamic nature of real-world data. Self-training methods overcome this limitation by fine-tuning the pre-trained LLMs using curated self-generated reasoning traces, thereby updating the model weights—unlike test-time methods where reasoning is induced via in-context learning without weight updates. This approach, first illustrated in Deepseek-R1 and fundamentally pioneered by STaR, has shown significant improvements in reasoning performance.

Assume that for a given problem, we have input-output pairs \((x_i, y_i)\), where \(x_i\) is a prompt and \(y_i\) is the correct answer, and let \(M_{PT}\) denote the pre-trained language model. The self-training process typically involves an arbitrary combination of the following techniques:
\begin{enumerate}
    \item \textbf{Supervised Fine-Tuning:} Fine-tune \(M_{PT}\) on the \((x_i, y_i)\) pairs using the causal language modeling loss to obtain \(M_{SFT}\).
    
    \item \textbf{Rejection Fine-Tuning:} For each prompt \(x_i\), generate a rationale (i.e. reasoning trace) \(\hat{r}_i\) and then an answer \(\hat{y}_i\) using either \(M_{PT}\) or \(M_{SFT}\). The generated triple \((x_i, \hat{r}_i, \hat{y}_i)\) is used for further fine-tuning only if \(\hat{y}_i = y_i\).
    
    \item \textbf{Preference Tuning:} Construct a preference dataset by generating two pairs \((x_i, \hat{r}_i, \hat{y}_i)\) and \((x_i, \hat{r}_j, \hat{y}_j)\) from the same model on input \(x_i\). If \(\hat{y}_j = y_i\) but \(\hat{y}_i \neq y_i\), then the pair \((x_i, \hat{r}_j, \hat{y}_j)\) is preferred. This preference data is then used—often via Direct Preference Optimization (DPO)—to fine-tune the model so that it favors generating reasoning traces that lead to the correct answer.
    
    \item \textbf{Reinforcement Learning:} Build a dataset of tuples \((x_i, \hat{r}_i, \hat{y}_i, R_i)\), where the reward \(R_i\) is 1 if \(\hat{y}_i = y_i\) and 0 or \(-1\) otherwise. The model is then updated using RL techniques to maximize the expected cumulative reward.
\end{enumerate}

STaR \cite{zelikman2022starbootstrappingreasoningreasoning} uses rejection fine-tuning on \(M_{PT}\) to produce a reasoning model \(M_{Reason}\). However, as rejection fine-tuning alone ignores negative training pairs, v-STaR \cite{hosseini2024vstartrainingverifiersselftaught} first obtains a model \(M_{RFT}\) via rejection fine-tuning and then applies DPO to yield \(M_{Reason}\). Other approaches, such as ReFT \cite{luong2024reftreasoningreinforcedfinetuning}, use reinforcement learning on \(M_{SFT}\) (obtained from \(M_{PT}\) via supervised fine-tuning), while the Self-Explore \cite{hwang-etal-2024-self} method begins with supervised fine-tuning to generate \(M_{SFT}\) and subsequently refines it through rejection-tuning and preference-tuning to obtain \(M_{Reason}\).
Figure \ref{fig:self-taughts} shows the overall training process of these four representative techniques for self-training.

\section{Challenges}

\begin{enumerate}
    \item \textbf{Challenges regarding Automation of Process Supervision Signals}: Developing PRMs (Process Reward Models) is hampered by the need for detailed process supervision signals—labels for every reasoning step—that currently depend on costly, labor-intensive human annotation \cite{lightman2023letsverifystepstep}. Automating this labeling process is essential for scalability and efficiency.
    
     \item \textbf{Computational Overhead and Overthinking}: While MCTS addresses issues with value-based networks in process supervision, it struggles with a vast search space and often overthinks, leading to unnecessary computational complexity and inefficiency \cite{luo2024improvemathematicalreasoninglanguage}. 
    
    \item \textbf{Expensive Step Level Preference Optimization}: Step level preference learning solves the issues DPO \cite{rafailov2024directpreferenceoptimizationlanguage} faces with reasoning task. However, it also presents significant challenges. Step level preference annotation is much more expensive compared to instance level annotation. It also requires fine-grained judgements that can lead to inconsistent and subjective labels.
    
    \item \textbf{Test-Time Compute Scaling depends on Robust Pre-training}: Test-Time Compute unlocks a model's best performance. This enables smaller models to outperform bigger models on easier questions \cite{snell2024scalingllmtesttimecompute}. However, a bottleneck arrives for more difficult questions. If the base model lacks robust pre-training, additional inference compute may not compensate for its deficiencies.
    
    \item \textbf{Test-Time Scaling Limitations}: Test-time scaling techniques such as Chain-of-Thought \cite{wei2023chainofthoughtpromptingelicitsreasoning} gained popularity because of its effectiveness and interpretable nature. However, \citet{wei2023chainofthoughtpromptingelicitsreasoning} also showed that Cot proved to be significantly effective only for LLMs with more than 100B parameters. For Smaller Language Models with less than 10B parameters, it also proved to be detrimental in certain cases.
\end{enumerate}

\section{Conclusion}

This survey provides a bird’s-eye view of the key techniques used to elucidate reasoning in language models. We deliberately kept the discussion accessible by omitting extensive mathematical derivations and detailed benchmark comparisons, which can often overwhelm newcomers to the field. Unlike many long surveys that are densely packed with technical details, our paper is relatively brief and focused, making it a less intimidating entry point for researchers new to the area. Our focus has been on presenting only the most essential information related to reasoning in LLMs. For those interested in a deeper exploration of specific aspects, we recommend consulting more comprehensive surveys such as \cite{kumar2025llmposttrainingdeepdive}.

\bibliography{custom}




\end{document}